# Understanding cyclists' perception of driverless vehicles through eye-tracking and interviews

Siri Hegna Berge*, Joost de Winter#, Dimitra Dodou#, Amir Pooyan Afghari†, Eleonora Papadimitriou†, Nagarjun Reddy*, Yongqi Dong*, Narayana Raju*, Haneen Farah*

*    Faculty of Civil Engineering and Geosciences, Delft University of Technology, The Netherlands
#   Faculty of Mechanical Engineering, Delft University of Technology, The Netherlands
†    Faculty of Technology, Policy and Management, Delft University of Technology, The Netherlands

Corresponding author: Joost de Winter, j.c.f.dewinter@tudelft.nl

**Highlights**
- This study used a Wizard-of-Oz method to assess whether cyclists can detect the absence of a driver in automated vehicles.
- We employed computer vision to analyse cyclists' gaze behaviour.
- Cyclists detected the absence or presence of a driver with 93% accuracy.
- Cyclists prefer to be informed about the vehicle's automated driving state.

**Abstract**
As automated vehicles (AVs) become increasingly popular, the question arises as to how cyclists will interact with such vehicles. This study investigated (1) whether cyclists spontaneously notice if a vehicle is driverless, (2) how well they perform a driver-detection task when explicitly instructed, and (3) how they carry out these tasks. Using a Wizard-of-Oz method, 37 participants cycled a designated route and encountered an AV multiple times in two experimental sessions. In Session 1, participants cycled the route uninstructed, while in Session 2, they were instructed to verbally report whether they detected the presence or absence of a driver. Additionally, we recorded participants' gaze behaviour with eye-tracking and their responses in post-session interviews. The interviews revealed that 30% of the cyclists spontaneously mentioned the absence of a driver (Session 1), and when instructed (Session 2), they detected the absence and presence of the driver with 93% accuracy. The eye-tracking data showed that cyclists looked more frequently and for longer at the vehicle in Session 2 compared to Session 1. Additionally, participants exhibited intermittent sampling of the vehicle, and they looked at the area in front of the vehicle when it was far away and towards the windshield region when it was closer. The post-session interviews also indicated that participants were curious, but felt safe, and reported a need to receive information about the AV's driving state. In conclusion, cyclists can detect the absence of a driver in the AV, and this detection may influence their perception of safety. Further research is needed to explore these findings in real-world traffic conditions.

## Introduction

Automated vehicles (AVs) are undergoing continuous development in terms of technology and capability (Liu et al., 2023; Parekh et al., 2022). The first commercial applications of robo-taxis are now available to the public (Public Utilities Commission, 2022; West, 2022), and advanced Level-2 AVs are driving on the roads (Nordhoff et al., 2023). Despite these technological strides, AVs grapple with imperfections in urban environments, where they may encounter pedestrians, cyclists, and other vulnerable road users (VRUs) (Matin & Dia, 2023; Sánchez et al., 2022). In particular, the interaction between AVs and cyclists requires more research due to the growing concerns over cycling safety (e.g., European Commission, 2023; SWOV, 2023).

A challenge in the development of AV technology is the issue of social communication. Both cyclists and drivers are known to engage in such communication, including eye contact, nodding, and hand gestures (Al-Taie et al., 2023; Färber, 2016; Li et al., 2023; Stanciu et al., 2018). However, in an AV, the driver is potentially inattentive or absent, leading to a potential breakdown in the communication process. This interaction void has been argued to be an obstacle to AV-VRU interaction (Rasouli & Tsotsos, 2020; Strömberg et al., 2018) and is cited as a reason for developing external human-machine interfaces (eHMI) mounted on the AV (Faas et al., 2020; Forke et al., 2021; Hensch et al., 2020; Rouchitsas & Alm, 2019; Wilbrink et al., 2021).

At the same time, a counter-narrative within the academic literature argues that the importance of social signals in traffic is being grossly overestimated (Lee et al., 2021; Moore et al., 2019). As shown in various studies (e.g., Pekkanen et al., 2022; Tian et al., 2023; Wang et al., 2010), crossing decisions of VRUs depend on implicit variables such as relative distance and speed. The AVs that have been developed so far do not incorporate eHMIs, barring a small number of exceptions (Ackerman, 2018; Mercedes-Benz, 2023). This raises the question of whether it is imperative to develop eHMIs or sensors that can detect the social signals of VRUs.

Consistent with this counter-narrative, a study using photographs found that VRUs often cannot detect the presence of a driver, particularly under conditions of darkness and windshield glare (AlAdawy et al., 2019). These findings are supported by observational studies in which pedestrians interacted with AVs in Wizard-of-Oz setups intended to create the illusion that the vehicle is driving automatically (Cefkin et al., 2019; Chen et al., 2020; Habibovic et al., 2018; Hensch et al., 2020; Moore et al., 2019; Rothenbücher et al., 2016; Sirkin et al., 2016). These studies showed that the absence of a driver behind the steering wheel was noticed in only about 80–90% of the encounters. Features of the AV, such as eHMIs, often go unnoticed as well, depending on the salience of the eHMI signal (Cefkin et al., 2019; Chen et al., 2020; Habibovic et al., 2018; Hensch et al., 2020). The existing Wizard-of-Oz studies aim to create a believable interaction with the AV through conspicuous signals. Pedestrians interacting with driverless vehicles appear to look longer out of curiosity (Li et al., 2020), sometimes accompanied by a certain playfulness (Currano et al., 2018), and as noted by Moore et al. (2019): "*without the props, it is likely few would have noticed the absence of a driver*" (p. 303).

Hensch et al. (2020) pointed out that not noticing a driver is not the same as seeing that the driver's seat is empty. They found that 79% of the participants correctly indicated no driver in the vehicle. However, when there was a driver in the vehicle, only 52% of the participants recognised this, presumably due to windshield glare. In other words, the results of Hensch et al. suggest that the methodology of this type of Wizard-of-Oz research merits improvement, by considering not only the possibility of misses (i.e., VRU fails to detect the presence of a driver) but also false positives (i.e., VRU reports the presence of a driver while there is no driver).

All Wizard-of-Oz studies cited above were conducted with pedestrians; there appears to be no research among cyclists, a category of VRUs who are likely to have more difficulty detecting drivers than pedestrians due to their higher speed, and, as found in earlier eye-tracking research (Pashkevich et al., 2022; Trefzger et al., 2021), their tendency to look at the road ahead, as they have to stabilise and navigate their bicycles. It is conceivable that cyclists, when faced with a visually challenging task like detecting a driver in an AV, will exhibit a coping strategy (also referred to as 'behavioural adaptation'), such as cycling more slowly.

The literature reveals a knowledge gap regarding whether VRUs, particularly cyclists, can detect the presence of a driver in the driver's seat. With the advent of AVs, cyclists will find themselves more often confronted with the task of estimating whether the AV is under manual or automatic operation, by gauging the presence of a driver. In light of this, the current article aims to address the following two research questions:

(1) Do cyclists notice the absence of a human driver when they pass an approaching AV?
(2) When explicitly asked to do so, how accurately can cyclists determine whether or not there is a human driver behind the steering wheel?

We also examined *how* cyclists performed these tasks, specifically:
(3) How do cyclists perform a driver detection task, in terms of eye movements and cycling speed adjustments?
(4) How do cyclists experience their interactions with AVs, particularly regarding the need for additional vehicle information and distinguishing between AVs and conventional vehicles?

Answering these research questions is important for determining whether modifications are needed in the design of AVs, for example regarding eHMIs that explicitly indicate whether the vehicle is being driven manually or automatically, or to assess whether cycling safety might be compromised when they try to make such an estimation.

A field experiment involving human participants was conducted to answer the research questions. In Session 1, participants repeatedly cycled past an approaching vehicle without receiving specific information. In Session 2, the participants were informed that the vehicle could drive automatically, and they were explicitly instructed to verbally report whether someone was behind the wheel. In our study, we did not use a real AV but instead used a hidden human driver. We used a variety of measurement methods, including verbal utterances during cycling, post-session semi-structured interviews, mobile eye-tracking, and determination of cycling speed.

## Methods

### Participants

The sample consisted of 37 participants (23 males and 14 females), ranging from 22 to 67 years of age ($M$ = 31.9, $SD$ = 11.0). The sample size of 37 was chosen based on practical and statistical considerations. Practically, it was the maximum achievable, as the study was conducted on weekends when the campus was relatively quiet. This was important because the temporary closure of the bicycle path, necessary for the experiment, had to be minimised. Statistically, a sample size of 37 was deemed sufficient to draw reasonably reliable conclusions about whether cyclists can notice drivers. For example, if 19 out of 37 participants detected the driver (51%), the 95% confidence interval would range from 34% to 68%, based on a binomial distribution. This level of accuracy was considered adequate to address Research Questions 1 and 2.

Regarding participants' highest educational degree completed, 16 (43%) had completed a bachelor's degree, 20 (54%) had a master's degree, and 1 held a degree from a Higher Technical School (a former institution in The Netherlands that provided vocational education in engineering). The participants reported cycling on average 4.75 hours a week ($SD$ = 3.36), and the majority reported "commuting or utilitarian" as their primary motivation for cycling (73%). Regarding the participants' vision, one participant reported "moderate difficulty" to the question "Do you have any difficulties (e.g., blurred vision) seeing or detecting objects while being on the road?", while 7 reported "a little difficulty"; the remaining 29 (78%) reported "no difficulty at all".

Participants were recruited using convenience sampling through flyers distributed on the TU Delft campus and via social media, as well as invitations through our personal and professional networks. The study was approved by the university's Human Research Ethics Committee (approval number 2300). All participants provided written informed consent.

### Experimental design

The experiment was of a within-subjects design consisting of two sessions with different instructions for the participants and two driver states. In each session, the participants cycled the same route four times, encountering a vehicle once per trial. In total, the participants thus passed the vehicle eight times. The average time between consecutive passages (three in Session 1, three in Session 2) across participants was $M$ = 135.1 s, with a standard deviation ($SD$) of 26.3 s ($n$ = 37).

In the literature, various Wizard-of-Oz methods have been used to investigate pedestrians' responses to AVs. These include a tinted windshield (Bindschädel et al., 2023; Chen et al., 2020; Lau et al., 2024), a dummy steering wheel (Habibovic et al., 2018), a mirror film (Liu & Hirayama, 2023; Liu et al., 2021), or the car being controlled from the passenger seat (Rodríguez Palmeiro, Van der Kint, Vissers, et al., 2018).

However, the most common option is that of a seat costume (Faas & Baumann, 2021; Fuest et al., 2018; Hensch et al., 2020; Joisten et al., 2020; Large et al., 2023; Li et al., 2020; Liang et al., 2016; Loew et al., 2022; Rothenbücher et al., 2016; Taima & Daimon, 2023; Wang et al., 2021). In our study, a concealed experimenter drove the vehicle in a seat costume during half of the encounters (see Figure 1). The costume was designed to simulate an AV by making the driver appear as an empty seat. It covered the driver's upper body and head with fabric shaped like a car seat. When in disguise, the driver gripped only the lower part of the steering wheel while operating the vehicle.

**Figure 1**
*Seat costume used in the experiment.*

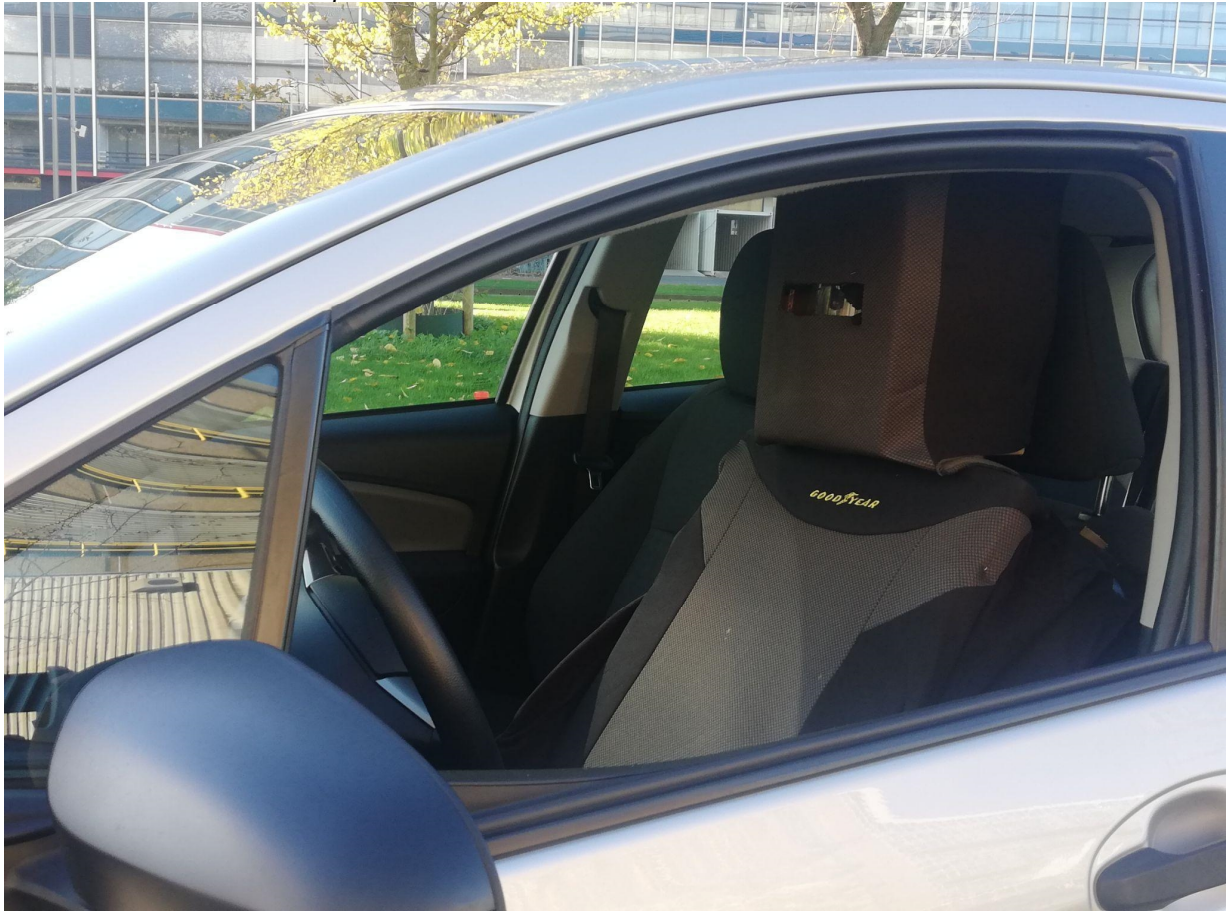

In Session 1, participants were neither led to believe nor informed that the vehicle could operate without a driver. In Session 2, participants were informed about the dual state of the vehicle, either human-driven or driven in automated mode. For Session 2, the participants were also instructed to verbally state either "Yes, driver" or "No driver" as soon as they could determine the presence or absence of a driver.

In two out of the four encounters in each session, the vehicle had no visible driver. The presence of the driver in each session followed one of four predefined sequences: No-Yes-Yes-No, Yes-No-No-Yes, Yes-Yes-No-No, or No-No-Yes-Yes. The sequence differed among participants but was consistent for the same participant across Sessions 1 and 2. This pattern was used to facilitate the role transition of the driver (i.e., experimenter).

**Procedure**
Data collection occurred over three weekends in November and December 2022. Collecting data during the weekend allowed for the closing of the test area with fences to minimise the interference of other cyclists and pedestrians and to ensure the safety of our participants. In addition, we had guards placed at potential entry points of the test area, preventing interference from outside road users and passersby.

Upon arrival, the participants received an information sheet and signed an informed consent form. They were equipped with Tobii 3 eye-tracking glasses, calibrated using a bullseye card held by an experimenter at a distance of about 0.8 m from the participant. The calibration was conducted before each of the two sessions. In sunny conditions, participants wore infrared-blocking tinted protective lenses (Tobii, 2022) attached to the eye-tracking glasses, and wore a baseball cap (Evans et al., 2012; Simpson et al., 2019) to minimise interference by sunlight.

*Session 1*
The participants received the instruction to cycle a 530-metre route around TU Delft in the Netherlands. They were told to cycle like they usually would. The participants used their own bikes during the experiment, except for two participants who used the bicycle of an experimenter. All participants used a conventional, non-electric bike. While cycling, the participants interacted with a grey 2018 Toyota Yaris. With its headlights on, it approached from the participant's left-hand side at an angle of approximately 160 degrees (Figure 2).

**Figure 2**
*Top-down view of the interaction scenario. The sign-holding experimenter was present in Session 2 only (overlay drawn on an image from Google Earth, 2023).*

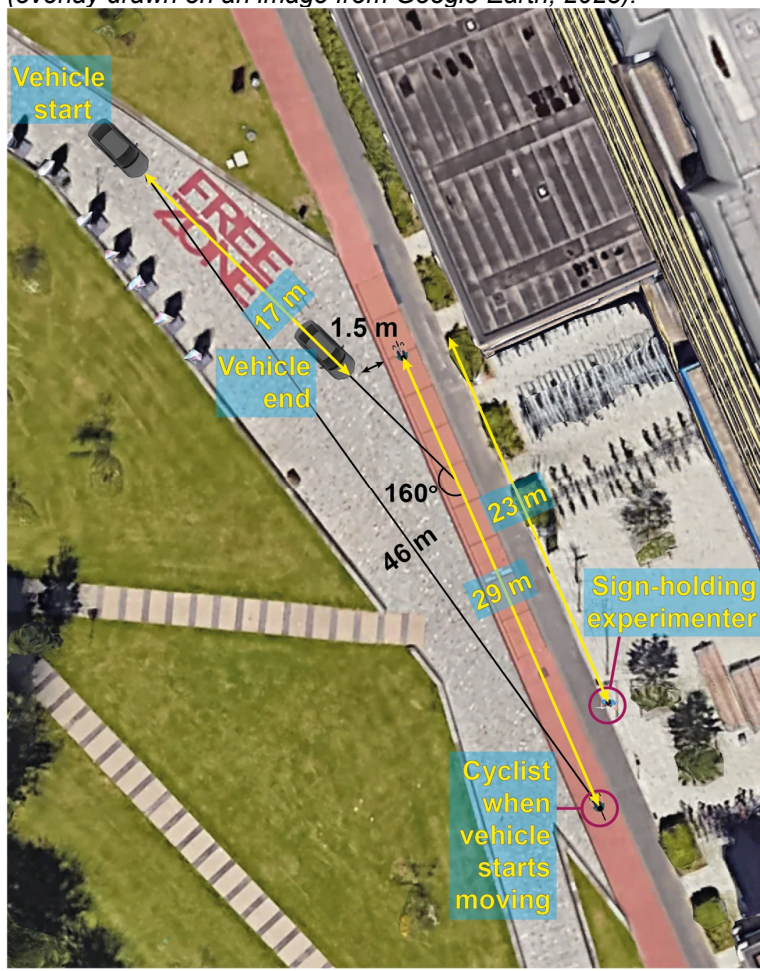

The driver proceeded straight ahead with a speed of about 10 km/h and stopped with the vehicle's left front edge approximately 1.5 metres from the edge of the bike path (see Figures 2 and 3). The speed and distance from the bike path were chosen based on pilot tests to create ambiguity about the traffic interaction. While driving in 'no driver' mode, the concealed driver placed their hands at the lower part of the steering wheel. In 'human-driven mode', the driver placed their hands at the top of the steering wheel and looked straight towards the cycling participant. In Session 1, the participants cycled the route four times consecutively before briefly being interviewed about their experience.

*Session 2*
At the end of the interview after Session 1, the participants were informed that the vehicle they were interacting with was driven either by a human or was in automated mode, as a driverless vehicle.

Session 2 involved cycling the same route with the same behaviour of the AV, but with additional tasks for the cyclist. During each of the four trials, the participants had to read two letters on a sign held by an experimenter on the right-hand side of the cycling path (Figure 3). We alternated between signs with the two-letter combinations of "HN", "NH", and "OD". The participants passed the vehicle after cycling a further 23 metres (an average of 5.4 s, depending on cycling speed). They indicated whether they could detect a driver in the vehicle by uttering "Yes, driver" or "No driver" at the earliest possible moment they could discern the driver's presence or absence. The microphone of the eye-tracker recorded their utterances.

We included the secondary task of reading letters before passing the vehicle to increase visual attention demands (see Figure 3). In this way, we prevented the possibility of cyclists prioritising the driver-detection task in an unrealistic manner, such as by focusing their attention solely on the vehicle. Additionally, performance on the letter-reading task served as a way to obtain a control measure of the participant's level of perception. Our analysis of the audio showed that the 34 participants performed the letter-reading task with very high accuracy, with 134 correct answers, 0 incorrect answers, and 2 trials in which a participant did not mention the letters at all.

After Session 2, the participants were interviewed a second time. At the end of the experiment, participants completed a questionnaire regarding their demographics and cycling habits. Finally, participants received a debriefing on how the vehicle was controlled. For participating, they received a voucher of €10 and a baseball cap.

**Eye-tracking data analysis**

*Pre-processing of eye-tracking data*
The eye-tracking data were processed according to Onkhar et al. (2024). First, the *x* and *y* gaze-point data were filtered with a moving median filter with 0.3-s time interval. The median filter ensures that high-frequency jitter is removed and that short-term gaps, for example, caused by blinking, are filled in. The remaining gaps in the data were filled with a linear interpolation. The eye-tracking data were stored at a frequency of 100 Hz, while the scene camera stored frames at 25 Hz. A gaze *x* and *y* point was calculated per video frame by averaging the four available values for each frame.

*Automatic identification of the vehicle in the scene camera*
In the analysis of mobile eye-tracking data, the question arises as to how these data should be analysed. One option involves manual annotation of whether the gaze point falls within an area of interest (AOI), either per fixation (Gruden et al., 2021; Vos et al., 2023) or frame-by-frame (Vansteenkiste et al., 2015). Aside from the fact that this method is subjective (Pfeiffer & Renner, 2014; Vos et al., 2023), it is also time-consuming (Meißner et al., 2019). Furthermore, cycling encompasses considerable head movement and variability in the bicycle's roll angle (Kovácsová et al., 2016). Therefore, a cyclist's field of vision is constantly in motion, and the definition of a 'fixation' (i.e., the eye focusing on a single point for a certain duration) may not be applicable (Vansteenkiste et al., 2015).

In light of these considerations, the decision was made to identify AOIs automatically, using a computer-vision method named YOLO (Redmon et al., 2016). YOLO accepts an image as input and outputs bounding boxes and associated object classes.

We used YOLOv4 (Bochkovskiy et al., 2020), pretrained on the COCO dataset (Lin et al., 2014; from sbairagy-MW, 2021). The COCO dataset comprises 80 object classes. However, we only used the 'car' class for our study. Each frame recorded by the eye-tracker scene camera was submitted to the YOLO algorithm, from which we extracted the bounding box surrounding the vehicle.

**Figure 3**
*Participant interacting with the experimental vehicle on the predetermined route (top) and an experimenter holding the sign for the letter-reading task (bottom).*

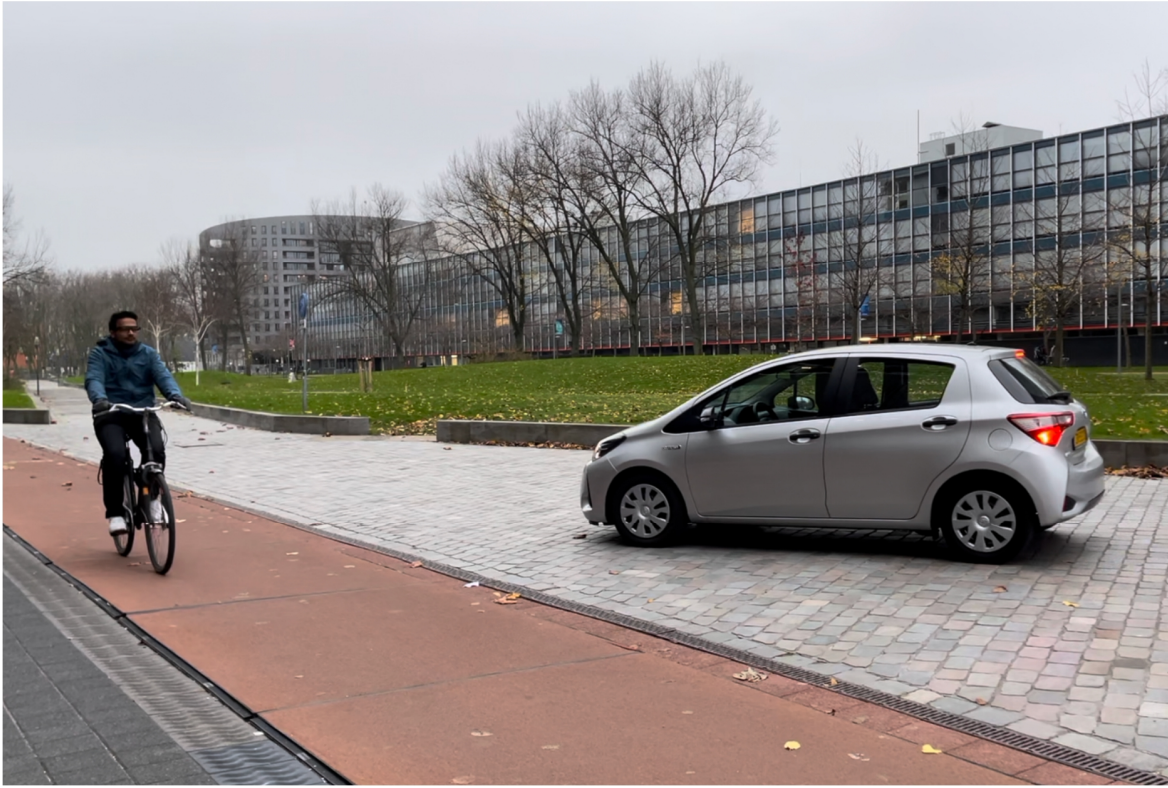

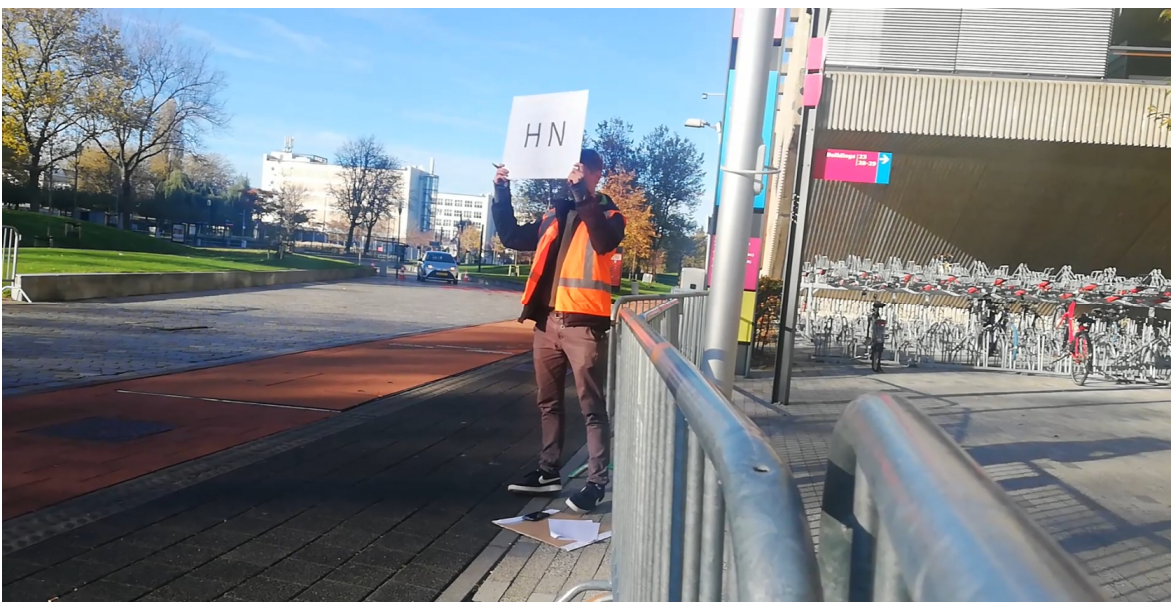

*Calculation of viewing angle*

We calculated, per video frame, the viewing angle between the gaze point and the bounding box surrounding the vehicle. The viewing angle describes the minimum angle the cyclist's eyes would have to rotate to look at the vehicle. The formula for viewing angle uses calibrated values of the eye-tracker concerning the number of pixels of the camera (1920×1080) and its focal length (Onkhar et al., 2024)[1].

Figure 4 illustrates the concept for a video frame, in which the participant would need to rotate their eyes by 33 degrees to look at the bounding box. If the gaze point fell within the bounding box, the viewing angle was defined as 0 degrees. It was assumed that the participant was looking at the vehicle if the angular distance was less than 4 degrees, based on Onkhar et al. (2021).

**Figure 4**
*A still frame from the video recorded by the eye-tracking glasses, with an overlay of the gaze point and a bounding box of the interacting vehicle.*

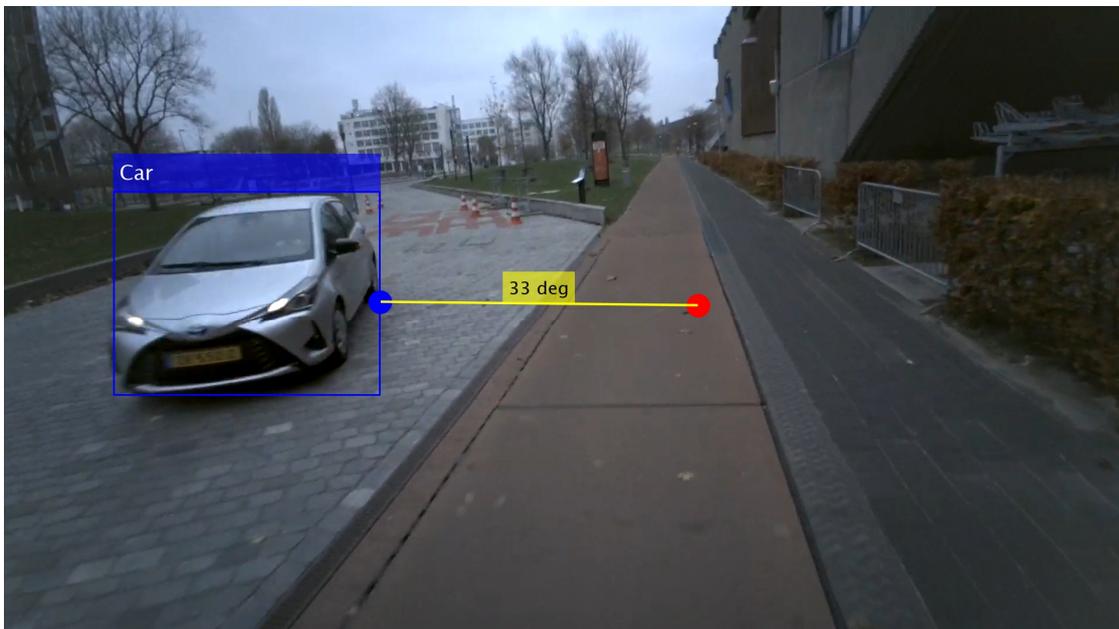

*Note.* The red circle indicates the participant's gaze point, and the blue box surrounds the vehicle.

Figure 5 illustrates the angular distance over a trial, i.e., a single interaction with the vehicle. The *x*-axis represents the elapsed time, where *t* = 0 s is the moment at which no portion of the vehicle was in view of the scene camera anymore. The *y*-axis shows the angular distance, with a value below 4 degrees being considered as the cyclist looking at the vehicle.

**Determining bicycle speed**

Cycling speed was determined manually: All videos were annotated to determine the elapsed time when passing nine landmarks surrounding the interaction with the vehicle. These landmarks were salient features, such as manholes on the cycling path and walkway entries located on the left side of the cycling path. In total, there were four landmarks prior to the AV and five landmarks after it (see Figure B1 in Appendix B). The distance between consecutive landmarks varied between 13.70 and 36.64 metres, and the total distance between the first and last landmarks was 182.6 metres.

---

[1] Note that we could also determine this distance in pixels. However, the use of an angle provides more generalisable and human-centric information than pixels, which depend on the camera resolution and camera field of view.

The mean speed between two subsequent landmarks was estimated as the distance between the two landmarks as measured in Google Maps divided by the time the participant took to cycle between them, where elapsed time was estimated from the corresponding number of frames in the video divided by the frame rate (see Kovácsová et al., 2019, for a similar approach).

**Figure 5**
*Angular distance between gaze point and vehicle for a single cyclist-AV encounter.*

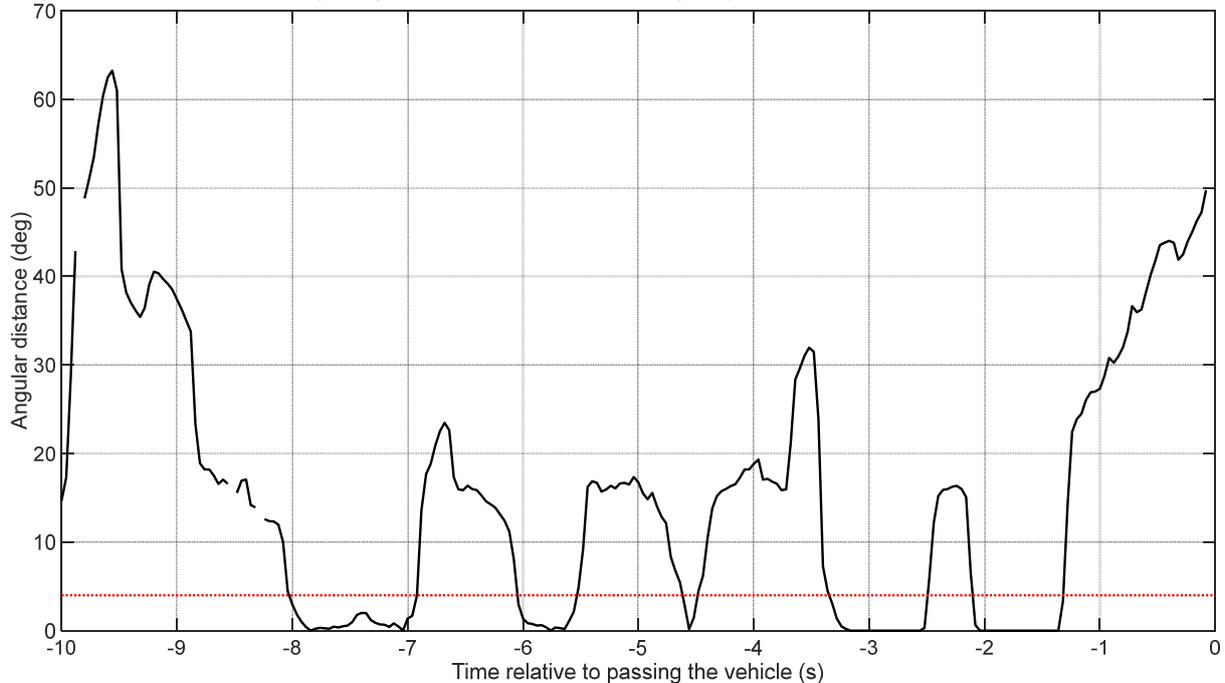

*Note.* The horizontal line is depicted at an angle of 4 degrees. A time of 0 s corresponds to the first video frame from the Tobii scene camera in which no part of the vehicle was visible anymore.

**Dependent measures**
We extracted the following dependent measures per trial:

- *Looking-at-vehicle.* Whether the participant was looking at the vehicle in the 10-s interval preceding the interaction (see Figure 5). Note that this measure is binary; it indicates whether, in the 10-second window, the vehicle was looked at, once or multiple times (1), or not at all (0).
- *Looking-at-vehicle-percentage.* How long the participant looked at the vehicle, defined as the percentage of time, determined over the 10-s interval.
- *Maximum speed*: The maximum speed from the eight distance intervals over which the cycling speed has been calculated.
- *Relative minimum speed*: The minimum speed from the eight distance intervals over which the cycling speed has been calculated, divided by the aforementioned maximum speed. This ratio indicates how much the participants had slowed down.

For Session 2, we additionally determined the following measure:
- *Driver detection accuracy.* Whether the participant correctly reported whether there was a driver in the vehicle. We manually extracted whether participants said "Yes, driver" or "No driver" from the eye-tracker's audio recordings. These utterances were compared with the ground truth, namely whether a human was behind the wheel in the respective trial. This analysis could yield 1 of 4 possibilities per trial: a true positive, a true negative, a false positive, or a false negative.

**Interview analysis**
The interview guides for Session 1 and Session 2 are shown in Tables 1 and 2.

**Table 1**
Guide for the interview after Session 1.

| # | Question |
|---|---|
| | *While cycling, you encountered a vehicle …* |
| Q1 | How did you interact with this vehicle? |
| Q2 | What did you look at? |
| | *[Probe for looking at certain areas of the car (e.g., windshield, hood, wheels), motion cues, searching for eye contact etc].* |
| Q3 | Did anything stand out? If so, what? |
| | *[Probe for whether they noticed anything different between the vehicles].* |
| Q4 | Did you interact differently with this vehicle as opposed to vehicles in daily life? How? Why? |

*Note.* The text segments in brackets indicate prompts or cues not always conveyed to the participants.

**Table 2**
Guide for the interview after Session 2.

| # | Question |
|---|---|
| | *While cycling, there was no driver in half of the times you encountered the vehicle ...* |
| Q5 | To what extent did the interactions affect your *[cycling]* behaviour? |
| | From 1 (not at all) to 5 (very much). |
| Q6 | How was your *[cycling]* behaviour affected? |
| Q7 | Did you feel safe? Please elaborate on your answer. |
| Q8 | Would you like to be able to distinguish the vehicle from a "normal vehicle"? |
| | *[e.g., a sign or marking that indicates automation].* |
| Q9 | Would you like additional communication from the vehicle? *[e.g., extra lights or a display]* |
| | Please elaborate on your answer. |
| | *[Probe for the preferred modality of communication: visual, auditory, haptic, connectivity, vehicle, infrastructure, bike, cyclist].* |
| Q10 | To what extent do you think the experiment setup was realistic? |
| | From 1 (not realistic) to 5 (very realistic). |
| Q11 | Is there anything else you would like to add? |

*Note.* The text segments in brackets indicate prompts or cues not always conveyed to the participants.

We recorded the interview audio with an Olympus VP-20 voice recorder. The average interview duration following Session 1 and Session 2 was 3 and 8 minutes, respectively. The audio was transcribed in clean verbatim by Amberscript (2023, May 1) and manually checked to correct the identification of the interviewer and interviewee.

The interview data were analysed with OpenAI's ChatGPT, using a prompting method presented by Tabone and De Winter (2023). Each question per participant was summarised using the ChatGPT API (model: gpt-4-0613). In this manner, we created 407 summaries (37 participants x 11 questions). An example prompt is provided in Table 3.

**Table 3**
Prompt for summarising a participant's transcript, with regard to one question (Q5).

*This is a transcript of an interview done with a participant who had cycled four rounds and interacted with a Wizard-of-Oz automated vehicle in each round. Before this session, the participant was told to look for and verbally indicate whether a driver was present.*
*Speaker 1 is the interviewer; Speaker 2 is the participant.*
*This is the transcript:*
*…*
*…*
*Based on this interview, what did the participant say about the extent to which the interactions affected their cycling behaviour? ; do not report numbers; only explanations*

As a validation check of the ChatGPT output, the interview data from the first 15 participants of Session 1 were analysed manually by one of the authors, creating brief summaries of each interview question (Q1–Q4) per participant. Another author compared these summaries with the transcripts and ChatGPT summaries. Compared to manual analysis, the summaries generated by ChatGPT contained more repetitive statements per participant across Q1–Q4, but captured details slightly better. We did not find any hallucinations in the ChatGPT output; see supplementary data for the comparison of ChatGPT and manual summaries.

The 37 ChatGPT summaries per question were subsequently summarised again for each question, creating meta-summaries. A prompt through which we achieved this, in this case for Q5, is displayed in Table 4. The meta-summaries are described in the results section, accompanied by illustrative quotes from the transcripts. The complete ChatGPT meta-summaries are provided in Appendix A.

**Table 4**
Prompt for summarising the summaries of all 37 participants, with regard to one question (Q5).

*Please summarise the above summaries into a meta-summary, capturing the views of all 37 participants, in four sentences. The output should capture the essence in an interesting and specific manner, in UK English.*
*Participant No. 1: …*
*Participant No. 2: …*

While ChatGPT is linguistically proficient, its numerical counting accuracy is debatable (Wang et al., 2024). To investigate how many participants noticed the absence of a driver during the interviews after Session 1, we manually analysed the transcripts, focusing on the responses from Q1 through Q4 (see Table 1). One researcher categorised participants into three groups:
1. Participants who mentioned the absence of a driver, e.g., "*The first round there was no one in the car*" (P6) and "*I noticed the second round that it looked like there was nobody in the car*" (P8).
2. Participants who referred to a 'driver', 'man', or 'person' inside the vehicle without mentioning their absence, e.g., "*I didn't look much at the driver*" (P35).
3. Participants who did not mention the notion of a driver at all in the transcripts, e.g., describing the speed and movement or physical appearance of the vehicle: "*Assess the speed and how it compares to my speed. (…) It's more [about] general movement*" (P25) and "*I just want to ride my bike. That's it. I don't care. I noticed that it's a white car*" (P13).

## Results
**Driver presence/absence recognition (Session 1)**
Following Session 1, the participants were interviewed. Out of the 37 participants, 11 correctly mentioned that (in some trials) the vehicle was devoid of a driver (Table 5). An additional 19 participants mentioned the notion of a driver but did *not* point out that there was no driver in the vehicle in some trials. The remaining 7 participants did not mention the notion of a driver in the interview.

**Table 5**
*The number and percentage of participants recognising driver presence or absence during the interview after Session 1*

| Driver presence/absence recognition | Number of participants | % |
|---|---|---|
| Mentioned that a driver was absent | 11 | 30% |
| Did not mention that a driver was absent | 19 | 51% |
| Did not mention the notion of a driver | 7 | 19% |

*Note. n* = 37.

**Accuracy of detection of driver presence and absence (Session 2)**
The confusion matrix of the driver detection task in Session 2 shows mostly accurate detections of the driver's presence or absence, with 63 true positives and 64 true negatives (Table 6). Conversely, incidences of false negatives (misses) and false positives were small (5 and 4, respectively). A total of 6 participants were responsible for the 9 mistakes. Note that, of the 37 participants, no eye-tracking camera and audio data were available for 3 participants in Session 2 due to a software error. Therefore, this analysis was based on 136 trials of 34 participants.

**Table 6**
*Confusion matrix depicting participants' detection of driver presence/absence, self-reported during Session 2 (total number of trials: 136)*

|  |  | **Ground truth** | |
|---|---|---|---|
|  |  | **Driver present** | **Driver absent** |
| **Participant** | "Yes, driver" | 63 | 4 |
|  | "No driver" | 5 | 64 |

**Looking at the vehicle (Sessions 1 & 2)**
As pointed out above, of the 37 participants, no eye-tracking data were available in Session 2 for three participants due to a software error. In addition, for one participant, the eye-tracking data from the first of four trials of Session 1 was not used because the vehicle did not drive due to an experimenter error.

Figure 6 depicts the percentage of participants looking at the vehicle as a function of time during the interaction. Note that a time of 0 s corresponds to the vehicle completely disappearing from the eye-tracker scene camera view. It can be seen that participants looked substantially more at the vehicle in Session 2 than in Session 1.

These findings are supported by quantifying the percentage of participants who looked at the vehicle at least once during a trial, i.e., the looking-at-vehicle measure. This measure was averaged across the 4 trials per session, resulting in a score with five possible values: 0%, 25%, 50%, 75%, or 100%. The average for Session 1 was 87.8%, while for Session 2 it was 94.9%. A Wilcoxon signed-rank test indicated that this difference was not statistically significant, $p = 0.375$ ($n = 34$). Note that the percentages are high, but still conservative, as eye-tracking data are imperfect. It is possible that a participant was looking at the vehicle, but the eye-tracker did not capture this as within 4 degrees of the bounding box.

**Figure 6**
*Percentage of trials in which participants looked at the vehicle, defined for each video frame across a 10-s window preceding the encounter.*

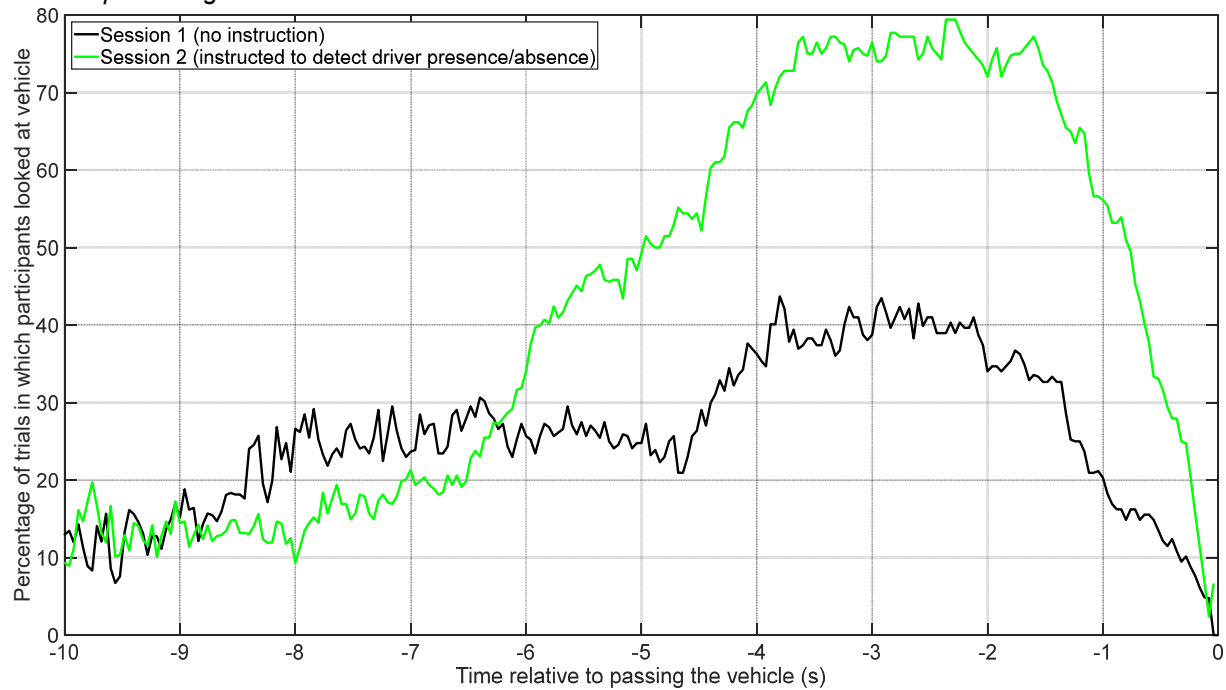

The percentage of time spent looking at the vehicle over the 10-s window was 27.4% (*SD* = 18.7%) in Session 1 versus 44.5% (*SD* = 19.4%) in Session 2. This difference was statistically significant according to a paired-samples *t*-test, *t*(33) = −4.06, *p* < 0.001.

**Gaze distribution**
The next question addressed is *how* the cyclists divided their attention while approaching the vehicle. To this end, all gaze data were partitioned into groups based on the distance between the cyclist and the vehicle, in 5-metre increments. The distance was estimated from the bounding box height in pixels. Only plausible bounding boxes were retained, with a width/height ratio between 1.00 and 1.90. Heatmaps were created from these partitioned data.

The heatmaps, depicted in Figures 7 and 8, show the average size of the bounding box in a green outline in consecutive order. The smaller the bounding box, the further away the vehicle is. The heatmaps were created by counting the number of gaze points in a grid of 20×20 pixels.

In Session 1 (Figure 7), a bimodal distribution in the viewing direction can be observed for all distances. That is, the cyclists frequently looked towards the vehicle but also frequently to the right of the vehicle, i.e., towards the bike path being ridden. The closer the cyclist approached the vehicle, the greater the horizontal distance between the two point-clouds. This can be explained by the fact that the vehicle and the bike path show an increasing angular separation as the vehicle gets closer. Another interesting phenomenon in Session 1 is that at far distances (Distance > 15 m), the cyclists did not look directly at the vehicle but lower than the vehicle. When the vehicle was closer, the cyclists looked at the vehicle, i.e., within the bounding box.

**Figure 7**
*Heatmaps of gaze distribution relative to the bounding box surrounding the vehicle, for different distances to the vehicle, for Session 1.*

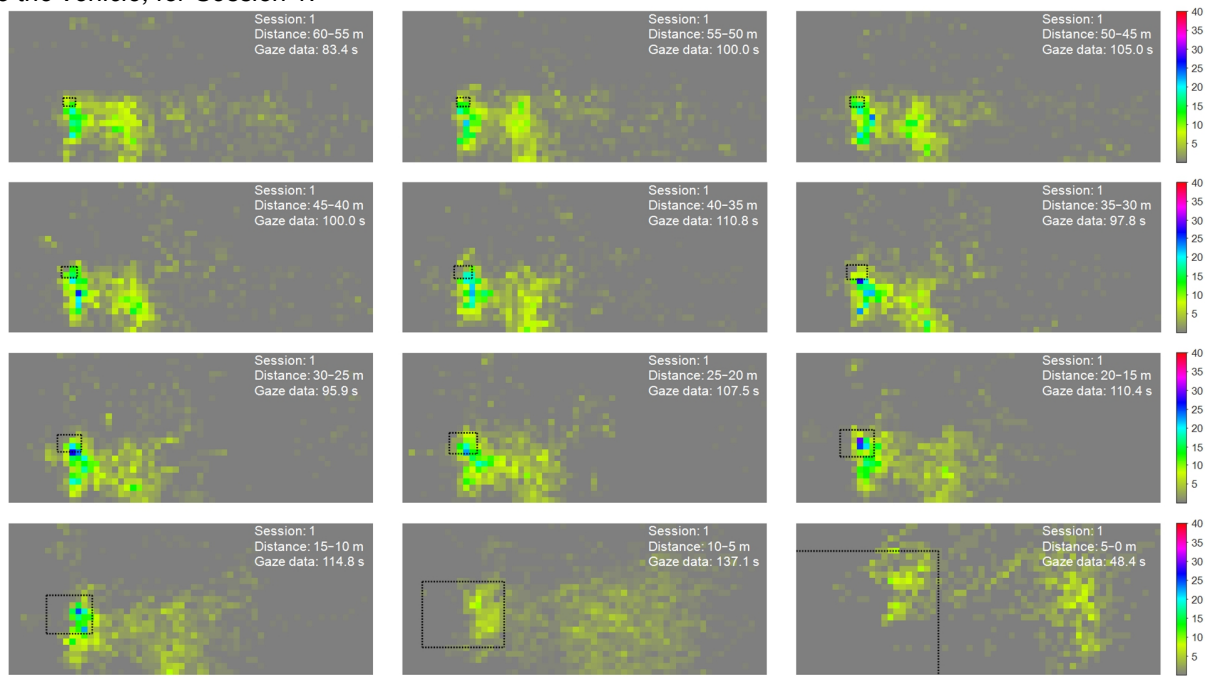

*Note.* The heatmaps consist of 20×20-pixel cells and are 1200 pixels wide and 500 pixels high. The sum of the depicted values equals 1000. The mean dimensions of the bounding boxes are presented by the dotted black rectangle.

The viewing distribution in Session 2 is substantially different from Session 1 (see Figures 7 and 8). It can be seen that the bimodal distribution at a far distance to the vehicle (Distance > 40 m) is more pronounced compared to Session 1, which can be explained by the fact that participants had the secondary task of reading letters on the sign held by the person positioned on the right-hand side of the cycling path. Shortly after (Distance < 40 m), participants focused their attention to a great extent on the vehicle itself. When the vehicle was close (e.g., Distance < 15 m), the difference from Session 1 is clearly noticeable. In Session 2, the gaze was concentrated around the location of the driver (top right corner of the bounding box). In contrast, in Session 1, it was much less directed into the bounding box and less clustered (yet still on the right-hand side of the vehicle, from the cyclist's perspective). In Figure 6, it was shown, and statistically substantiated, that in Session 2, participants overall looked more at the vehicle than in Session 1. A similar pattern is visible in Figures 7 and 8, but now as a function of the distance between the cyclist participant and the vehicle, and thus less diluted by the different cycling speeds participants adopted. For example, for the distance interval 50–45 m, participants looked more at the vehicle in Session 1 ($M$ = 25.7, $SD$ = 25.8) than in Session 2 ($M$ = 10.3, $SD$ = 15.4), $t(33)$ = 4.75, $p$ < 0.001 (explainable due to the letter task). When the vehicle was closer, for example for the distance interval 10–5 m, participants looked less at the vehicle in Session 1 ($M$ = 25.5, $SD$ = 27.6) than in Session 2 ($M$ = 65.0, $SD$ = 31.2), $t(33)$ = −6.17, $p$ < 0.001.

**Figure 8**
*Heatmaps of gaze distribution relative to the bounding box surrounding the vehicle, for different distances to the vehicle, for Session 2.*

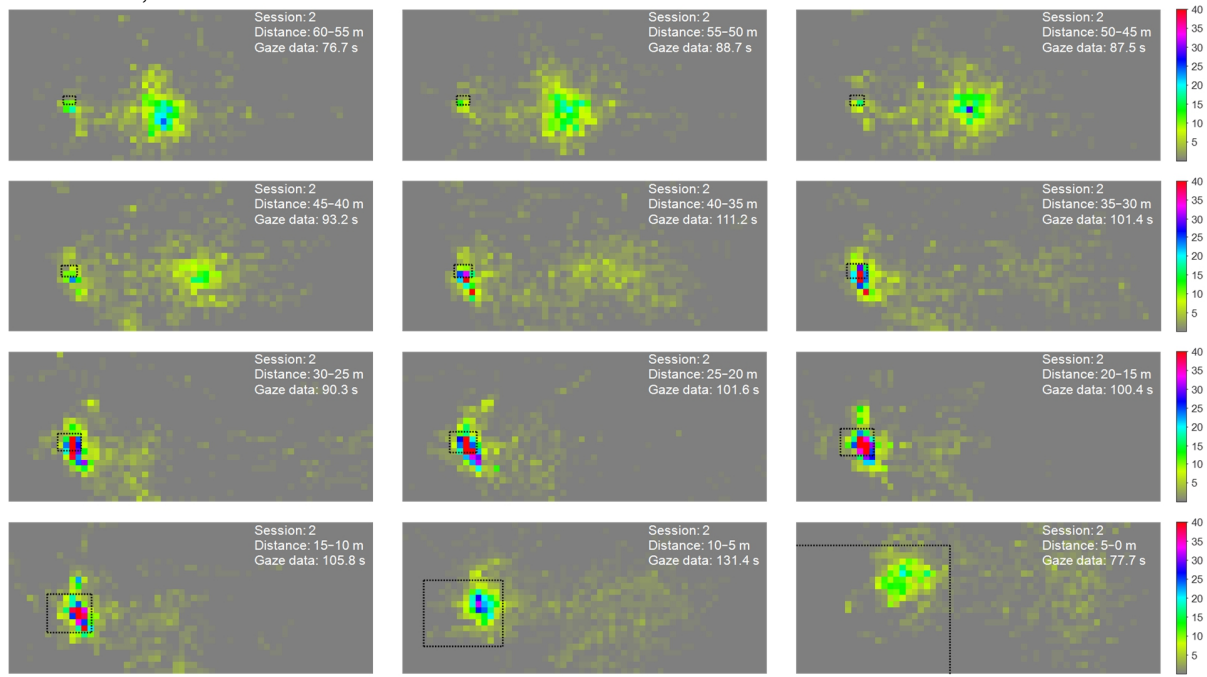

*Note.* The heatmaps consist of 20×20-pixel cells and are 1200 pixels wide and 500 pixels high. The sum of the depicted values equals 1000. The mean dimensions of the bounding boxes are presented by the dotted black rectangle.

**Cycling speed (Sessions 1 and 2)**
The maximum speed at which participants cycled was not significantly different in Session 1 ($M$ = 18.1 km/h, $SD$ = 3.3 km/h) versus Session 2 ($M$ = 18.1 km/h, $SD$ = 3.0 km/h), $t(33)$ = −0.53, $p$ = 0.598. Figure 9 (left) illustrates that the speed correlated strongly between Sessions 1 and 2 ($r$ = 0.96).

The relative minimum speed was significantly lower in Session 2 ($M$ = 0.837, $SD$ = 0.052) than in Session 1 ($M$ = 0.864, $SD$ = 0.044), $t(33)$ = 3.67, $p$ < 0.001. This could indicate that participants in Session 2 slowed down in an attempt to better identify whether there was someone behind the wheel. These effects are shown in Figure 9 (right).

There also appeared to be a learning effect where the relative minimum speed in the first trial of a session was consistently low (Session 1: 0.822, 0.877, 0.871, 0.883; Session 2: 0.781, 0.850, 0.855, 0.863, for Trials 1, 2, 3, and 4, respectively).

**Figure 9**
Left: Maximum speed (average of 4 trials) in Session 2 versus Session 1. Right: Relative minimum speed (average of 4 trials) in Session 2 versus Session 1.

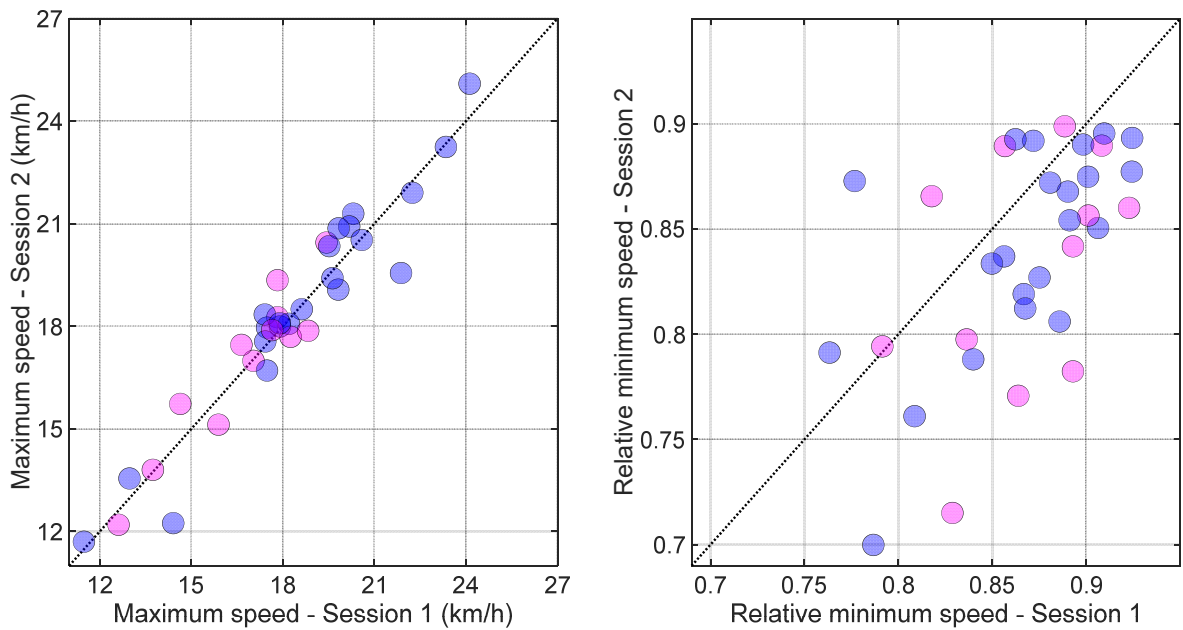

*Note.* Each marker represents a participant (blue: male, magenta: female).

**Interview data**

*Session 1*
Table A1 in Appendix A provides the full ChatGPT meta-summaries of the interviews after Session 1. In short, the interviews revealed that participants interacted with the vehicle by observing its speed and adjusting their cycling speed accordingly (Q1). Participants frequently focused on the front of the vehicle and tried to identify the driver. They often observed the vehicle's speed and direction but rarely mentioned looking for specific cues or making eye contact (Q2).

> *"I started looking and sort of, you know, calculating whether it would be a hazard for me or not. Judging by its speed and how far into the cycling area it was".* (P39)

While some participants noticed variations in vehicle speed and driver presence, and a few spotted unusual vehicle movement, many did not observe differences between the four trials (Q3). Participants interacted somewhat differently with the vehicle than normal due to its slower speed and the controlled experimental environment, and they generally felt safe due to its predictable behaviour (Q4). Some participants reported confusion, but also increased caution and curiosity in response to the driverless condition (Q4).

> *"First two times: no driver in there. So I was like, what is happening here? So I was pretty confused about it, but how did I interact with it? It was going that slow, so I didn't feel like we were going to collide with each other".* (P11)

> *"I tend to stare at the vehicle longer when there's no driver in it because it's a little bit strange. (…) The first time I was a little bit shocked and then, a little bit curious about how this car works (…)".* (P15)

> *"At first, I didn't notice, but when I just passed it and I saw, 'Oh, there are no people'. So, I just turn my head back [to] see it. (…) Just out of curiosity".* (P16)

***Session 2***
Tables A2–A4 in Appendix A contain the ChatGPT meta-summaries of the interviews after Session 2. These summaries revealed that the interactions with the AV often increased caution in cyclists, with driver absence influencing some participants' behaviour, similar to the participants noticing that the vehicle sometimes was driverless in Session 1. Participants also said they trusted or treated automation technology like any other vehicle (Q5). Many felt safer with a driver than without, but the trust in the AV increased over time (Q6). While the task of identifying the driver affected their eye movements and cycling speed, most still felt safe due to the vehicle's slow speed and the controlled setting (Q6, Q7).

> *"I saw that the speed was very moderate. So even if I didn't see the driver, I thought, okay, this vehicle for sure won't hit me. (…) looking at the speed, I estimated that [it] would have not hit me".* (P9)

> *"It's not on the street, it's in the campus. So, in this kind of situation, I feel safer. And I know I'm participating in the experiment".* (P23)

The majority of participants expressed a preference for being able to differentiate AVs from traditional vehicles.

> *"(…) I think in more busy traffic it would be nice to know how unpredictable a person can be or a road user can be. So yeah, I think it would be nice to know whether there is a driver driving or not".* (P36)

> *"I would want to know whether a car is currently being driven by some sort of robotic system. I mean, anything could go wrong. So I'd rather be able to know that if I'm going to look inside the car, probably I'm going to find people distracted. So, I shouldn't shock myself because that's happening; I would know. I would make the distinction that that car is [automated]".* (P39)

However, a minority of participants did not need such distinctions as long as the AV was reliable (Q8). Visual communication methods such as lights, signs, or displays were the most favoured, but some also mentioned auditory communication, especially in complicated or high-speed scenarios (Q9). A few participants also suggested haptic feedback or connectivity, although these methods were considered secondary to visual signals.

> *"I think a [visual] sign would do for me. And of course, you can also imagine that you wear a wearable, and it interacts with the car and it communicates through haptic communication".* (P6)

Most participants preferred the vehicle to be the source of communication. A minority felt that additional communication was unnecessary or expressed concerns about information overload (Q9).

> *"In traffic, there's already a lot going on. It can also be kind of information overload"* (P25).

Some participants suggested that the experiment needed more elements like road users and obstacles. Also, certain elements in the experiment, such as the experimenter holding the letters and the presence of pedestrians at a specific corner, were questioned by some participants. Furthermore, some participants suggested varying the vehicle's position, speed, and direction, indicating that the existing conditions might have felt too predictable (Q11).

**Discussion**

While some research suggests that road users rely on social cues from drivers (Färber, 2016; Li et al., 2023; Stanciu et al., 2018), other studies indicate that VRUs rarely pick up on such social cues (Lee et al., 2021; Moore et al., 2019). The objective of this experimental study was to determine whether cyclists can notice if a vehicle is driverless (Research Question 1) and whether they can detect the presence or absence of a driver in the driver's seat (Research Question 2). Additionally, we examined how cyclists performed these tasks using eye-tracking and speed measurements (Research Question 3) and how they experienced their interaction with the AV (Research Question 4).

Our experiment with 37 cyclists found the following: in a post-session interview, 30% (11 out of 37) of the cyclists spontaneously noted that the vehicle had no driver (Research Question 1), and when asked to do so, the cyclists could detect whether someone was behind the wheel with 93% accuracy (Research Question 2). In previous 'ghost driver' studies in which VRUs reported 'no driver', it was impossible to ascertain whether this was because the VRU did not notice the driver due to inattention or windshield glare, or because they could actually discern there was no driver. In our study, we accounted for this issue by measuring false negatives (misses) and false positives, discovering an even distribution between the two, which indicates minimal response bias. These findings have implications for research on the interaction between VRUs and AVs. It is often argued that vehicle kinematics (i.e., implicit communication) are more important than eye contact or other forms of communication from the driver, or their substitution by eHMIs (e.g., Moore et al., 2019; see Introduction). Our results support this claim, as cyclists often failed to notice the absence of a driver. However, the results also show that when cyclists were tasked to detect the driver, they did notice the absence of the driver. This suggests that signals from the driver could be used by VRUs when they feel the need for additional communication (e.g., when implicit communication is insufficient).

Apart from investigating cyclists' ability to recognise the presence or absence of a driver, we used eye-tracking to examine how cyclists performed this task (Research Question 3). The results showed that the cyclists' attention was bimodally distributed, with cyclists looking either at the path ahead or at the vehicle. These findings differ from pedestrian gaze behaviour (De Winter et al., 2021) but are consistent with previous research on gaze behaviour among cyclists (Pashkevich et al., 2022; Trefzger et al., 2021): Cycling is a steering task that requires intermittent attention. Our results resonate with classic research on car driving (Godthelp, 1988; Senders et al., 1967), in which occlusion techniques were used to examine how drivers steer their vehicles while their vision was intermittently occluded.

When the vehicle was far away (> 15 m), our participants tended to look at the area in front of the vehicle. As the vehicle got closer, the cyclists predominantly looked at the vehicle itself, particularly at the windshield. We could not find similar findings in the literature, except for Dey et al. (2019), who investigated the gaze behaviour of pedestrians as a vehicle approached them. A potential reason for the cyclists' focus in front of the vehicle could be that they were trying to anticipate its future position. Other forms of anticipatory viewing have been observed among VRUs. For example, studies among pedestrians (De Winter et al., 2021) and motorcyclists (Aupetit et al., 2015; Papakostopoulos et al., 2020) have found that they do not necessarily focus on the nearest vehicle but rather on vehicles further away, paying attention to distinct features such as turned wheels, to better predict the path of these vehicles.

This study also found that cyclists mainly looked at the right side of the vehicle. One possible explanation is that the driver was positioned on this side of the vehicle. An alternative explanation is that this strategy requires smaller eye movements. According to a visual attention model by Wickens (2008), people's gaze behaviour is partly determined by 'effort', and humans are not inclined to make larger eye and head movements than needed. Future research, for example, on cars with the steering wheel on the left versus the right, should provide further clarification.

In our study, we observed behavioural adaptation: participants in Session 2 slowed down, regardless of the driver's presence. Reducing cycling speed affords the cyclist additional time to detect the driver, and may enhance the quality of perception, since it reduces optical flow and vibrations. Additionally, speed reduction improves safety and enables cyclists to focus on the vehicle instead of primarily on the bicycle path. However, slower cycling speeds can reduce the bicycle's stability, implying that cycling too slowly can also be unsafe (Schwab & Meijaard, 2013; Sharp, 2008). Previous research into cyclist-AV interaction has used video clips (Thellman et al., 2023; Vlakveld et al., 2020) and still photographs (Bazilinskyy et al., 2023; Ngwu et al., 2022), or asked participants to imagine themselves interacting with an AV (Sahaï et al., 2022), approaches which neglect the adaptive nature of cycling. In our study, the individual differences in cycling speed were highly consistent from Session 1 to Session 2 and showed significant gender differences. These findings align with previous studies on vehicle driving (De Winter & Happee, 2012; Groeger, 2001), with the physical strength of cycling as additional explanatory power for gender differences in cycling speed (Kovácsová et al., 2016).

The post-session interviews (Research Question 4) revealed a novelty effect, with participants demonstrating curiosity and alertness in their encounters with the driverless vehicle, a pattern also observed in pedestrian-AV interaction (Li et al., 2020). However, some participants expressed confidence in the AV's safe and reliable behaviour. Previous research suggests that the absence of a driver lowers perceived safety, but this effect diminishes when the vehicle has an eHMI (Bazilinskyy et al., 2021; Faas & Baumann, 2021). Other research showed that perceived safety depends on the presumed reliability of the AV. For example, online research by Rodríguez Palmeiro, Van der Kint, Hagenzieker et al. (2018) found that cyclists felt more certain that the AV had detected them and would be inclined to proceed when they were informed that the AV's sensors were reliable. Similarly, research among pedestrians found that a vehicle that was crawling forward slowly (not unlike our study) was interpreted as poor human driving by some pedestrians but as advanced automation by others (Bazilinskyy et al., 2021). Onkhar et al. (2022) found that a driver's eye contact influenced pedestrian crossing intentions but only when vehicle motion cues were ambiguous. Similarly, Al-Taie et al. (2023) showed that cyclists use gestures and on-vehicle signals in less regulated situations. In summary, from past research, it appears that responses of VRUs to driverless vehicles are context-sensitive and dependent on their beliefs about the reliability of the AV (see also Parkin et al., 2023; Thellman et al., 2023). Longer-term studies where participants encounter both traditional and driverless AVs are required to make conclusive statements about how the lack of a driver behind the wheel influences cycling experience and behaviour.

Our research may have implications for the development of eHMIs that inform road users about an AV's actions (e.g., Bindschädel et al., 2023; Colley et al., 2021; Forke et al., 2021), its awareness of them (e.g., Block et al., 2023; Eisele & Petzoldt, 2022; Epke et al., 2021), or its automated driving mode (e.g., Daimon et al., 2021; Joisten et al., 2020). Results from the interviews following Session 2 indicated that most participants would like to receive such information, especially through visual eHMIs on the AV. However, eHMI developers should consider that cyclists apply intermittent visual sampling, which implies that eHMIs should be readable in a single glance. Participants were more sceptical about solutions requiring connectivity, such as bike-mounted devices. Hence, the present findings provide a sobering perspective on the many on-bike warning systems described in the literature (Berge et al., 2023).

At the same time, AV developers should be responsible for improving the detection of cyclists and other VRUs (Berge et al., 2022). AVs still struggle to predict road-user behaviour (Rahmati et al., 2020; Song et al., 2022), and as a result, they often drive conservatively or brake unnecessarily (Aittoniemi, 2022; Yang et al., 2023). Sometimes referred to as 'phantom braking' (Berge et al., 2024; Linja et al., 2022; Moscoso et al., 2021; NHTSA, 2022), this phenomenon may explain why AVs are frequently rear-ended (Kutela et al., 2022; Ma et al., 2022; Petrović et al., 2020). Cyclists can also trigger such braking events in automated shuttles by overtaking them (De Ceunynck et al., 2022; Pelikan, 2021; Pokorny et al., 2021). Moreover, studies indicate that test drivers frequently take over control out of safety necessity (Boggs et al., 2020; Houseal et al., 2022) or embarrassment (Lindgren et al., 2022; Nordhoff & De Winter, 2023). Therefore, advancements in sensor technology and artificial intelligence (AI) are important for enhancing AV perception capabilities and increasing overall safety.

**Limitations**
One limitation of this research is that despite its real-world setting, it was still staged. In the interviews, participants questioned the vehicle's presence in a campus area restricted to vehicles. Berge et al. (2024) highlighted 20 key scenarios for future cyclist-AV interaction studies. Our research corresponds with the 'perpendicular vehicle' scenario. Further research should explore scenarios like right-turning vehicles, left turns, parked vehicle situations, and scenarios requiring negotiation.

Another limitation is that Sessions 1 and 2 were not counterbalanced but always presented in the same order. This was necessary to measure the spontaneous detection of the driver (Session 1) before explicitly making cyclists aware of the task of detecting the driver (Session 2). The downside of this is that some learning effects or other temporal effects could have occurred, including fatigue or boredom. Future research could use an alternative study design, such as a between-subjects design. Another limitation of the experiment was that it involved varying lighting conditions because it was conducted across three full days in November and December. While weaker sun intensity typical of these months reduces problems with infrared radiation affecting the eye-tracker (Tobii, 2020; Vos et al., 2023), the low sun angle could increase glare, possibly impairing driver visibility.

The participants were primarily recruited from a technical university, which limits the representativeness of our findings. Additionally, although all participants reported engaging in weekly cycling, the cycling behaviour of several international participants appeared to differ from typical Dutch cycling behaviour: They displayed caution, characterised by their avoidance of the vehicle and a slower cycling pace. It would be interesting to conduct further research on the interaction between cyclists and (automated) vehicles in countries where cycling is less common or where cycling infrastructure is less available.

It is unknown how many participants believed the vehicle was driving automatically. Previous Wizard-of-Oz research showed believability percentages ranging from 97 to 100% (Faas & Baumann, 2021; Habibovic et al., 2018; Joisten et al., 2020), but also 60% to 88% (Currano et al., 2018; Faas & Baumann, 2019; Hensch et al., 2020; Large et al., 2023; Li et al., 2020; Moore et al., 2019; Rothenbücher et al., 2016), and 40% (Rodríguez Palmeiro, Van der Kint, Vissers, et al., 2018). These percentages appear to be context-dependent, for example, whether a modern-looking vehicle was used. In our study, our impression was that the participants generally believed that the vehicle was driving automatically, as noted in participant responses during the debriefing.

A final limitation involves our use of ChatGPT for generating interview summaries. Similar approaches have been previously used, with seemingly valid results (Patel & Lam, 2023; Tabone & De Winter, 2023). However, ChatGPT's usage of terms such as "some participants" or "many participants" is somewhat ambiguous. Therefore, it is recommended that a summary produced by ChatGPT should be used as a supplement to quotations and the manual extraction of counts, as we have done in our study.

## Conclusion

While cyclists often did not spontaneously mention the absence of a driver in a post-cycling interview, they were highly capable of detecting the presence/absence of a driver when specifically instructed to do so. Cyclists slowed down when trying to detect the presence of a driver. Furthermore, they focused either on the path ahead or towards the vehicle itself, indicating that cycling is a task that requires intermittent attention. When the vehicle was still far away, cyclists looked at the area in front of the vehicle rather than at the vehicle itself. Finally, participants reported increased caution and curiosity when encountering a driverless vehicle but also expressed trust in the AV's safe behaviour. They wanted clear communication from AVs, such as visual signals or signs, to differentiate AVs from traditional vehicles. Participants preferred the AV as the source of communication rather than relying on external devices or infrastructure.

Future research on cyclist-AV interaction should be conducted in more diverse traffic scenarios. Furthermore, it would be beneficial to investigate the perceptions and behaviours of a more diverse sample of cyclists. Finally, we recommend further exploring the use of computer vision for analysing eye-tracking data and large language models for analysing interview data; these approaches may offer promising opportunities for future research.

## Acknowledgements
The authors would like to acknowledge the seed funding received from the TU Delft Safety and Security Institute for the GAZETOAV project, which facilitated carrying out the field test of this study. Berge and De Winter have received funding from the European Union's Horizon 2020 research and innovation programme under the Marie Skłodowska Curie grant agreement 860410, the Institute of Transport Economics: Norwegian Centre for Transport Research, and The Research Council of Norway. Farah, Reddy, Dong, and Raju have received funding from the Applied and Technical Sciences (TTW), a subdomain of the Dutch

Institute for Scientific Research (NWO), through the project Safe and Efficient Operation of Automated and Human-Driven Vehicles in Mixed Traffic (SAMEN) under Contract 17187.

**Data availability**

Data and MATLAB scripts that reproduce the analyses in this paper are available at https://doi.org/10.4121/ee1abac8-bfc9-4a1b-9a45-29f1bc461eb

**Declarations of interest:** None.**References**

Ackerman, E. (2018, May 10). Drive.ai launches robot car pilot in Texas with a focus on humans. IEEE Spectrum. https://spectrum.ieee.org/driveai-launches-dallas-robot-car-pilot-with-a-focus-on-humans

Aittoniemi, E. (2022). Evidence on impacts of automated vehicles on traffic flow efficiency and emissions: Systematic review. *IET Intelligent Transport Systems*, *16*, 1306–1327. https://doi.org/10.1049/itr2.12219

AlAdawy, D., Glazer, M., Terwilliger, J., Schmidt, H., Domeyer, J., Mehler, B., Reimer, B., & Fridman, L. (2019). Eye contact between pedestrians and drivers. *Proceedings of the Tenth International Driving Symposium on Human Factors in Driver Assessment, Training and Vehicle Design*, Santa Fe, NM, 301–307. https://core.ac.uk/download/pdf/286570036.pdf

Al-Taie, A., Abdrabou, Y., Macdonald, S. A., Pollick, F., & Brewster, S. A. (2023). Keep it real: Investigating driver-cyclist interaction in real-world traffic. *Proceedings of the 2023 CHI Conference on Human Factors in Computing Systems*, Hamburg, Germany. https://doi.org/10.1145/3544548.3581049

Aupetit, S., Espié, S., & Bouaziz, S. (2015). Naturalistic study of riders' behaviour in lane-splitting situations. *Cognition, Technology & Work*, *17*, 301–313. https://doi.org/10.1007/s10111-014-0293-z

Bazilinskyy, P., Dodou, D., Eisma, Y. B., Vlakveld, W., & De Winter, J. (2023). Blinded windows and empty driver seats: The effects of automated vehicle characteristics on cyclists' decision-making. *IET Intelligent Transport Systems*, *17*, 72–84. https://doi.org/10.1049/itr2.12235

Bazilinskyy, P., Sakuma, T., & De Winter, J. (2021). What driving style makes pedestrians think a passing vehicle is driving automatically? *Applied Ergonomics*, *95*, 103428. https://doi.org/10.1016/j.apergo.2021.103428

Berge, S. H., De Winter, J., Cleij, D., & Hagenzieker, M. (2024). Triangulating the future: Developing scenarios of cyclist-automated vehicle interactions from literature, expert perspectives, and survey data. *Transportation Research Interdisciplinary Perspectives*, *23*, 100986. https://doi.org/10.1016/j.trip.2023.100986

Berge, S. H., De Winter, J., & Hagenzieker, M. (2023). Support systems for cyclists in automated traffic: A review and future outlook. *Applied Ergonomics*, *111*, 104043. https://doi.org/10.1016/j.apergo.2023.104043

Berge, S. H., Hagenzieker, M., Farah, H., & De Winter, J. (2022). Do cyclists need HMIs in future automated traffic? An interview study. *Transportation Research Part F: Traffic Psychology and Behaviour*, *84*, 33–52. https://doi.org/10.1016/j.trf.2021.11.013

Bindschädel, J., Weimann, P., & Kiesel, A. (2023). Using eHMI, acoustic signal, and pitch motion to communicate the intention of automated vehicles to pedestrians: A Wizard of Oz study. *Transportation Research Part F: Traffic Psychology and Behaviour*, *97*, 59–72. https://doi.org/10.1016/j.trf.2023.06.017

Block, A., Lee, S., Pandya, A., & Schmitt, P. (2023). I see you! Design factors for supporting pedestrian-AV interaction at crosswalks. *Companion of the 2023 ACM/IEEE International Conference on Human-Robot Interaction*, Stockholm, Sweden, 364–368. https://doi.org/10.1145/3568294.3580107

Bochkovskiy, A., Wang, C.-Y., & Liao, H.-Y. M. (2020). *YOLOv4: Optimal speed and accuracy of object detection.* arXiv. https://doi.org/10.48550/arXiv.2004.10934

Boggs, A. M., Arvin, R., & Khattak, A. J. (2020). Exploring the who, what, when, where, and why of automated vehicle disengagements. *Accident Analysis & Prevention*, *136*, 105406. https://doi.org/10.1016/j.aap.2019.105406

Cefkin, M., Zhang, J., Stayton, E., & Vinkhuyzen, E. (2019). Multi-methods research to examine external HMI for highly automated vehicles. In H. Krömker (Ed.), *HCI in mobility, transport, and automotive systems* (pp. 46–64). Springer. https://doi.org/10.1007/978-3-030-22666-4_4

Chen, H., Cohen, R., Dautenhahn, K., Law, E., & Czarnecki, K. (2020). Autonomous vehicle visual signals for pedestrians: Experiments and design recommendations. *Proceedings of the IEEE Intelligent Vehicles Symposium*, Las Vegas, USA, 1819–1826. https://doi.org/10.1109/IV47402.2020.930462820

**Appendix A.** ChatGPT output of the meta-summaries of the interview data

**Table A1**
*ChatGPT output for post-Session 1 interviews*

*Q1: How did they interact with the vehicle?*

*Participants interacted with the vehicle by observing its speed, direction, and the presence of a driver. They adjusted their cycling behaviour based on these observations, with some slowing down or altering their path to avoid potential collisions. Many participants noted changes in the vehicle's behaviour across the rounds, with some noticing the vehicle moved earlier or slower in later rounds. The presence or absence of a driver in the vehicle influenced some participants' reactions, with some feeling more cautious or curious when they realised the vehicle was driverless.*

*Q2: What did they look at?*

*Participants' attention was primarily drawn to the front of the vehicle, with many focusing on the headlights, speed, and direction. Some tried to identify the driver or make eye contact, but this was often hindered by factors such as dark glass or personal vision issues. A few participants noted the vehicle's behaviour, such as whether it would stop, and some observed the vehicle's colour, size, and brand. However, there was little mention of specific motion cues, and many did not notice whether there was a driver in the vehicle.*

*Q3: Did anything stand out? If so, what?*

*Participants' experiences with the automated vehicle varied, with some noting differences in speed, presence of a driver, and the vehicle's behaviour across different rounds. A few participants noticed the absence of a driver in some rounds, leading to feelings of curiosity, confusion, and increased caution. Others focused on the vehicle's speed, direction, and their distance from it, with some noting changes in these aspects across rounds. However, a significant number of participants did not notice any standout features or differences between the vehicles, treating the automated vehicle similarly to regular cars they encounter in daily life.*

*Q4: Did they interact differently with this vehicle as opposed to vehicles in daily life? How?*

*Participants' interactions with automated vehicles varied, with some treating them as they would regular vehicles, while others exhibited heightened caution. Factors influencing these interactions included the vehicle's speed, the controlled experimental environment, and the absence of a driver. Some participants felt safer due to the predictability of the automated vehicle, while others were more cautious due to the unfamiliarity and perceived unpredictability of the vehicle. The presence or absence of a driver did not significantly affect most participants' interactions, with many focusing more on the vehicle's behaviour and their own safety.*

**Table A2**
*ChatGPT output for post-Session 2 interview questions about interaction behaviour and safety*

### Q5: To what extent did the interaction affect their cycling behaviour?

The interactions with automated vehicles affected the cycling behaviour of the participants to varying degrees. Some participants felt the need to be more cautious, often slowing down to identify the presence of a driver, while others felt safe and unaffected due to the controlled nature of the experiment and the slow speed of the vehicle. The task of identifying the presence of a driver was a common focus, with some participants changing their cycling strategies to better observe the vehicle. However, a few participants expressed trust in the technology and did not feel the need to differentiate between human-driven and automated vehicles, suggesting that their cycling behaviour would remain consistent regardless of the presence of a driver.

### Q6: Describe how their cycling behaviour was affected.

The presence of automated vehicles affected the cycling behaviour of most participants, with many expressing a need to predict the vehicle's motion and identify whether a driver was present. This often led to increased caution and changes in speed or route. However, the majority felt safe due to the slow speed of the vehicles and some became more confident as they familiarised themselves with the vehicle's behaviour. Many participants expressed a desire for automated vehicles to be distinguishable from regular vehicles and for some form of communication about the vehicle's intentions.

### Q7: Did they feel safe? Elaborate.

The majority of the 37 participants felt safe interacting with automated vehicles during the experiment, attributing their comfort to factors such as the slow speed of the vehicle, trust in the technology, and the controlled experimental environment. Some participants expressed a desire for clear communication from autonomous vehicles, such as visual signs indicating their intentions or their driverless status, to enhance safety. A few participants felt less safe or had mixed feelings, citing concerns about real-world scenarios, the need for increased vigilance, and the unsettling experience of a car moving towards them without a driver. Overall, while participants generally felt safe, there were suggestions for improvements and some reservations about interacting with driverless vehicles outside of a controlled experiment.

**Table A3**
*ChatGPT output for post-Session 2 interview questions about cyclists' information needs*

### Q8: Would they like to be able to distinguish the vehicle from a normal vehicle? If so, how?

The majority of the 37 participants expressed a desire to distinguish automated vehicles from traditional ones, suggesting various methods such as signs, markings, or unique vehicle designs. A smaller group felt no need for such distinction, citing reasons such as trust in the system's safety, potential distraction, or a preference for treating all vehicles the same. Some participants were undecided or expressed nuanced views, suggesting that while distinction might not be necessary, some form of visual communication from the vehicle could be beneficial. Overall, the need for clear identification of automated vehicles was linked to factors such as safety, predictability of vehicle behaviour, and accountability in case of accidents.

### Q9: Would the participant like additional communication from the vehicle? If so, what kind? Describe placement, colour, and modality of communication if possible.

The majority of the 37 participants expressed a desire for additional communication from autonomous vehicles, with a strong preference for visual signals, such as lights or signs, directly from the vehicle itself. These signals could indicate the vehicle's intentions or confirm the cyclist's detection. A few participants suggested the use of auditory signals or haptic feedback, but these were generally seen as secondary to visual cues. However, a small number of participants felt that additional communication was unnecessary, preferring to rely on their own visual perception and existing traffic rules.

**Table A4**
*ChatGPT output for post-Session 2 interview wrap-up*

### Q10: How realistic did they find the experiment setup?

Participants had mixed views on the realism of the experimental setup involving automated vehicles and cyclists. Some found it very realistic, citing the behaviour of the automated vehicle and the environment as closely mirroring real-world conditions. However, others felt it lacked realism due to factors such as the absence of other road users, the predictability of the vehicle's behaviour, and the knowledge that they were in a controlled experiment. Suggestions for improvement included adding more complexity and unpredictability to the experiment, such as varying vehicle speeds, including more road users, and conducting the experiment in unfamiliar or more typical cycling environments.

### Q11: Anything else?

Participants in the study had varied responses, with many suggesting ways to increase the realism of the experiment. These included varying the vehicle's position, adding more people and bicycles to the environment, and introducing more moving objects or obstacles. Some participants found it challenging to identify whether there was a driver in the vehicle due to factors such as glass reflection and signage. Others questioned the purpose of certain elements of the experiment, such as the presence of pedestrians and the person holding letters, while a few expressed interest in the study's focus and potential implications for cycling speed.

**Appendix B.** Additional figures

**Figure B1**
*Position of the nine landmarks used to determine bicycle speed (overlay drawn on an image from Google Earth).*

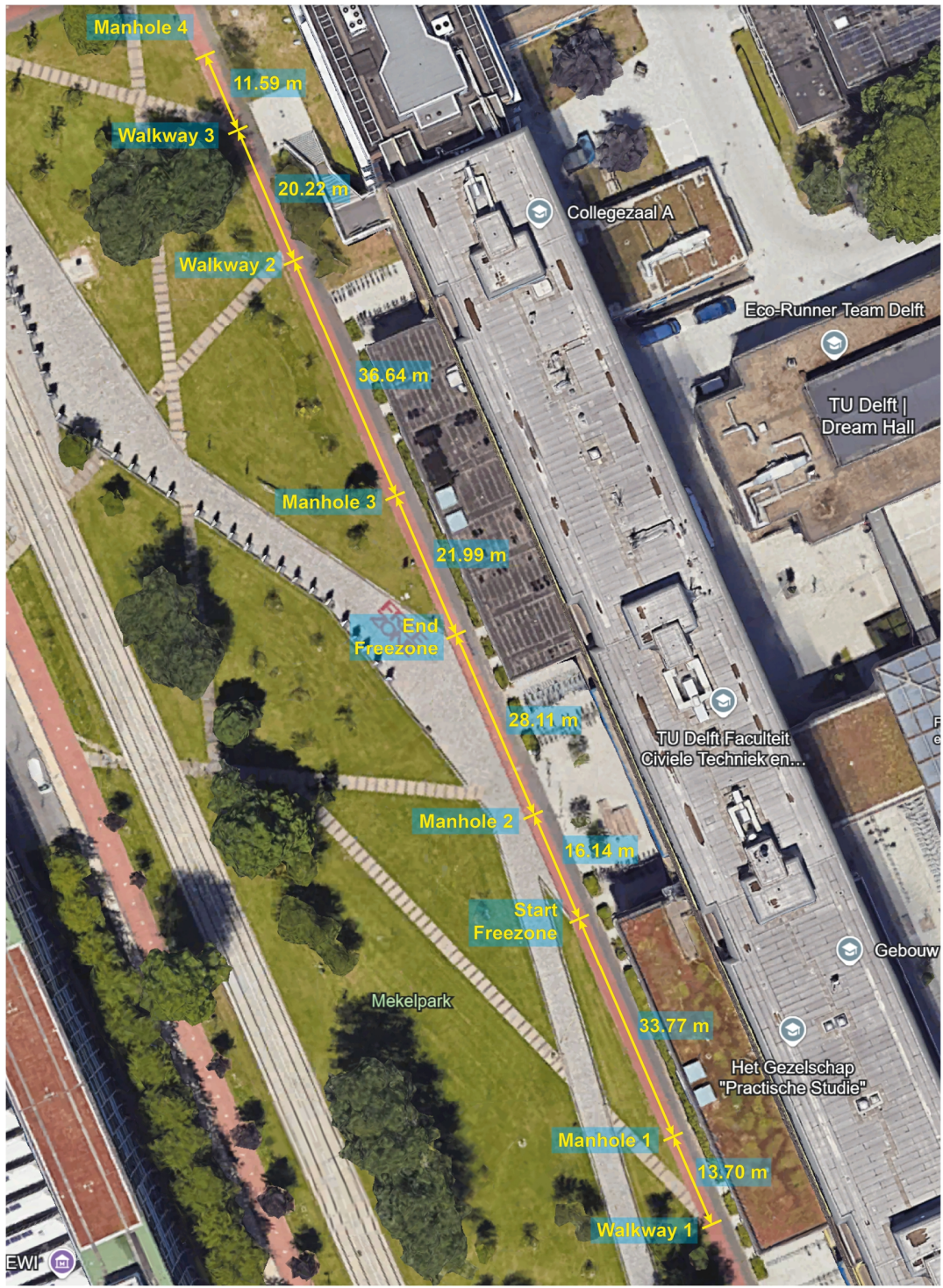

**Figure B2**
*Percentage of trials in which the YOLO algorithm was able to detect the vehicle, as a function of elapsed time in the trial. It can be inferred that at farther distances, the YOLO algorithm was less able to detect the vehicle.*

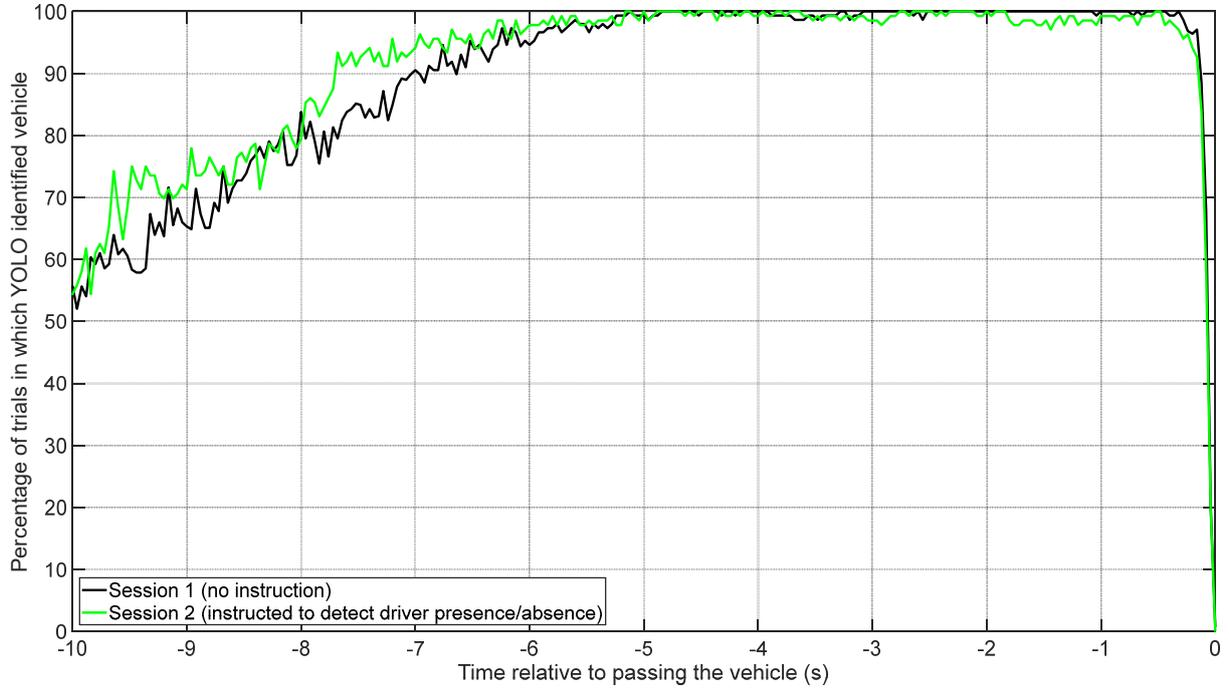